\documentclass[10pt,conference]{IEEEtran} 

\usepackage{color}
\usepackage{balance}
\usepackage{soul} 
\usepackage[spanish, english]{babel} 
\usepackage{url}
\usepackage[utf8]{inputenc}
\usepackage{enumitem} 

\usepackage{float}
\usepackage{pgfplots}
\usepackage{times}
\usepackage{balance}
\usepackage{graphicx}
\usepackage{booktabs} 
\usepackage{multirow} 
\usepackage{framed}   
\usepackage{amsmath}
\usepackage{pgfplots} 
\usepackage{blindtext}
\usetikzlibrary{patterns} 

\usepackage{stackengine} 
\usepackage{filecontents} 

\usepackage{amsmath}
\usepackage{algorithm}
\usepackage[noend]{algpseudocode}

\newcommand{\ToolURL}{\url{https://tacticvolatility.github.io/}} 

\usepackage{xspace} 

\newcommand{\eg}{\emph{e.g.,}\xspace}

\newcommand{\etal}{\emph{et~al.}\xspace} 
\newcommand{\descStep}[2]{\noindent \textbf{#1: } #2}

\author{\IEEEauthorblockN{Jeffrey Palmerino, Qi Yu, Travis Desell and Daniel E. Krutz} 
\IEEEauthorblockA{
B. Thomas Golisano College of Computing and Information Sciences\\
Rochester Institute of Technology, Rochester, NY, USA\\
Email: \{jrp3143, qyuvks, tjdvse, dxkvse\}@rit.edu}
}

\begin{document}









\title{Improving the Decision-Making Process of Self-Adaptive Systems by Accounting for Tactic Volatility}

\maketitle

\begin{abstract}

When self-adaptive systems encounter changes within their surrounding environments, they enact \textit{tactics} to perform necessary adaptations. For example, a self-adaptive cloud-based system may have a tactic that initiates additional computing resources when response time thresholds are surpassed, or there may be a tactic to activate a specific security measure when an intrusion is detected. In real-world environments, these tactics frequently experience \textit{tactic volatility} 
which is variable behavior during the execution of the tactic.

Unfortunately, current self-adaptive approaches do not account for tactic volatility in their decision-making processes, and merely assume that tactics do not experience volatility. This limitation creates uncertainty in the decision-making process and may adversely impact the system's ability to effectively and efficiently adapt. Additionally, many processes do not properly account for volatility that may effect the system's Service Level Agreement (SLA). This can limit the system's ability to act proactively, especially when utilizing tactics that contain latency.

To address the challenge of sufficiently accounting for tactic volatility, we propose a \emph{Tactic Volatility Aware} (TVA) solution. Using Multiple Regression Analysis (MRA), TVA enables self-adaptive systems to accurately estimate the cost and time required to execute tactics. TVA also utilizes \emph{Autoregressive Integrated Moving Average} (ARIMA) for time series forecasting, allowing the system to proactively maintain specifications. 

\end{abstract}


\section{Introduction}

The world is increasingly relying upon autonomous, self-adaptive systems that have the ability to function independently without human interaction. Examples of these self-adaptive systems include self-driving cars, medical devices, and many common Internet of Things (IoT) devices. Many self-adaptive processes utilize a closed-loop control mechanism that monitors the system's state and its surrounding environment. Furthermore, these mechanisms also determine if the system should be altered to perform any necessary adaptations~\cite{kephart2003vision, Brun:2009:ESS:1573856.1573860, salehie2009self, Dobson:2006:SAC:1186778.1186782}. These self-adaptive approaches typically rely upon a set of \emph{adaptation tactics} to make necessary changes~\cite{moreno2017adaptation, Schmerl2014ArchitecturebasedSC, 4221625, Huber:2014:MRA:2583044.2583093, Garlan2009}. Example tactics include the provisioning of an additional virtual machine in a web farm when the workload reaches a specific threshold, or reducing non-essential functionality on an autonomous Unmanned Aerial Vehicle (UAV) when battery levels are low.

Tactics frequently experience latency, which is the amount of time from when a tactic is invoked until its effect on the system is realized~\cite{moreno2017adaptation, Moreno:2018:FED:3208359.3149180, Moreno:2017:CMP:3105503.3105511, 7573126}. Examples of tactic latency include a cloud-based system requiring more than a minute to update certain firmware nodes~\cite{maatta2010program, anaya2014prediction}, or a cyber-physical system requiring one minute to re-activate GPS signals~\cite{liu2012energy, moreno2015proactive}. Tactics also frequently have a cost associated with them, which may be in the form of energy, monetary or other resource costs necessary for execution~\cite{Jung:2009:CAE:1813355.1813367,4220924,CostBenefitAnalysis}. Examples of tactic cost include the required energy for moving a physical component in a UAV or the monetary cost of using a remote third-party resource for computational tasks.

Both tactic latency and cost are likely to be first-class concerns in the decision-making process, as they can directly impact if and when a tactic is executed~\cite{moreno2017adaptation, Moreno:2017:CMP:3105503.3105511, 7573126}. Improperly accounting for tactic latency can lead to situations where tactics are begun too early or too late, or are not available when needed~\cite{moreno2017adaptation, Moreno:2018:FED:3208359.3149180}. Additionally, improperly accounting for tactic cost can result in the selection of a tactic that is more expensive than a tantamount, less costly alternative~\cite{CostBenefitAnalysis}. Therefore, it is imperative that self-adaptive systems properly account for both the latency and cost of tactics.

Real-world systems will frequently encounter \emph{tactic volatility}, which is any rapid or unpredictable change that exists within the attributes of a tactic. For example, both tactic latency and cost may be highly volatile depending on the system's surrounding environment. A tactic of transmitting data could take longer than expected due to network congestion, or a tactic of moving a physical component in a UAV could be more expensive due to mechanical problems.

Unfortunately, state-of-the-art decision-making processes in self-adaptive systems do not account for tactic volatility. This limitation can be highly problematic, adversely impacting the decision-making process in several ways. For example, a system may execute a tactic too late to be effective if it assumes that the tactic will always take two seconds to conduct, when in reality it has been observed to consistently take longer.

A \emph{Service Level Agreement} (SLA) helps to define system objectives such as keeping a value under a specific threshold, along with rewards and penalties for meeting defined objectives~\cite{moreno2017adaptation}. A challenge for purely reactive processes is that the system may not adhere to objectives defined in the SLA if tactics are expected to experience latency. For example, the SLA for a cloud-based self-adaptive system may define a reward for operating under a response time threshold. However, tactics such as adding servers to reduce response times will likely experience varying levels of latency. To perform optimally, the system should act proactively by beginning these tactics \emph{before} the additional resources are required so that the resources are available when needed~\cite{7573126,camara2014stochastic}. Thus, a mechanism that enables the system to accurately forecast tactic latency and cost values is highly beneficial to the self-adaptive process.





To address the limitations of current self-adaptive processes in properly accounting for tactic volatility, we propose a \emph{Tactic Volatility Aware} (TVA) solution. Through the use of a Multiple Regression Analysis (MRA) model, TVA enables the self-adaptive system to learn from previously experienced tactic volatility and make accurate estimates of how the tactic will behave in the future. TVA also includes an \emph{Autoregressive Integrated Moving Average} (ARIMA) component to enable self-adaptive systems to proactively begin adaptations according to their SLA specifications. 

To summarize, this work makes the following contributions:
\begin{enumerate}[noitemsep]

    \item \textbf{Problem Demonstration: }Using simulations, we demonstrate that accounting for tactic volatility is essential in self-adaptive systems, especially when the system is known to have unpredictable behavior.

    \item \textbf{Concept: }To the best of our knowledge, our TVA approach is the first process that accounts for tactic volatility. Existing processes merely consider tactics to have static values, whereas our TVA approach uses run time predictions of latency and cost to handle tactic volatility.

    \item \textbf{Experiments: }Our experiments demonstrate the positive impact that TVA has on the decision-making process. These experiments are conducted using real-world data, and thus provide additional confidence in our findings.


    \item \textbf{Tool and Dataset: }Our VolAtiLity EmulaTor (VALET) tool includes data generated from a physical system. This enables the evaluation of our TVA process and provides other researchers with a foundational dataset that they may use to support their own research. This tool and dataset are available on the project website: \ToolURL


\end{enumerate}

The rest of the paper is organized as follows. Section~\ref{sec: problemdefinition} motivates our research using examples of tactic volatility. Section~\ref{sec:proposedTechniqueThesis} defines our TVA solution to effectively address tactic volatility. Section~\ref{sec: VALETTool} describes \emph{VALET}, our tactic volatility dataset. Section~\ref{sec: Evaluation} describes our systematic experimental evaluation of our proposed process. Section~\ref{sec: relatedworks} describes related works, while Section~\ref{sec: threats} describes threats and future work. Section~\ref{sec: conclusion} concludes our work.

\section{Problem Definition} 
\label{sec: problemdefinition}

\emph{Tactic latency volatility} and \emph{tactic cost volatility} are the primary motivators for addressing a system's SLA with proactive processes. If a system is purely reactive and does not properly anticipate future system changes in regards to the specifications defined in the SLA, it will be limited by its inability to provide support for proactive tactic implementation. 

\subsection{Tactic Cost Volatility}

Examples of tactic cost vary widely and are largely domain specific, however tactic cost is frequently a primary concern in the decision-making process~\cite{CostBenefitAnalysis}. Possible examples of tactic cost include the energy necessary to move a mechanical component in a physical device, the monetary cost to utilize a resource, or the number of computations necessary to perform a tactic. Estimating tactic cost will likely be a first-class concern for the system, especially if there are defined cost thresholds, resource limitations, or if the system merely has the goal of attaining maximum utility at the lowest cost~\cite{moreno2017adaptation}. Unfortunately, despite the cost volatility that many real-world systems are likely to encounter, existing self-adaptive processes that account for cost consider it to be a static value that will encounter no volatility~\cite{Moreno:2018:FED:3208359.3149180, 7573126, moreno2015proactive}. Accounting for tactic cost volatility is imperative for several reasons:

\begin{enumerate}[noitemsep, nosep]

    \item \descStep{Cost may be a primary consideration when selecting between multiple tactic options}{When the system has multiple tactic options that it may choose from, the cost of the tactic may be a determining factor when selecting between multiple, otherwise tantamount options.}
    
    
    \item \descStep{Determine if the cost exceeds reward}{In selecting which tactic(s) to execute, the system will frequently calculate the expected reward and cost for performing the tactic. If the cost exceeds the reward, then it may not be optimal for the system to execute the tactic.}
    
    \item \descStep{The estimated cost may impact the system's ability to execute concurrent or subsequent tactics}{The system may possess a finite amount of a resource, with one example of being battery power. Assume that a system has 10 units of battery power remaining. Processes that consider cost to be a static value may define the energy usage of tactic $a=4$, and for tactic $b=5$. Therefore the system will assume that it has the ability to execute both tactics concurrently or sequentially. However if the actual energy usage of each tactic is $a=6$ and $b=7$, in reality the system will not have the ability to fulfill the amount of energy necessary for the execution of both tactics either concurrently or sequentially.}

\end{enumerate}

\subsection{Tactic Latency Volatility}

The amount of time required to implement a tactic is known as \emph{tactic latency}~\cite{camara2016analyzing}, which in real-world systems, can be highly volatile. 
For example, the precise amount of time a system needs to transmit a file across a network may fluctuate due to varying amounts of network traffic. Previous works have shown that latency aware self-adaptive processes offer several advantages over traditional, non-aware techniques~\cite{camara2014stochastic, moreno2017adaptation, Moreno:2018:FED:3208359.3149180}. However many state-of-the-art self-adaptive processes still consider tactic latency to be a predetermined and static value. Systems lack the capability to learn to adaptively accommodate for variability in tactic latency. Due to its large impact on the decisions that a system should make, accounting for tactic latency volatility is imperative for several reasons:

\begin{enumerate}[noitemsep,nosep]
    \item \descStep{Knowing when to begin the execution of a tactic}{A priority for many self-adaptive systems is to ensure that tactics are ready when needed. Therefore, if a tactic has expected latency, then the system will need to proactively begin its execution so that it is available when needed. However, implementing a tactic proactively will typically have additional costs involved, so proactive adaptation must be done with consideration to available resources.}
    
    





    \item \descStep{Determine when to augment a slow tactic with a faster tactic}{Accurate tactic latency knowledge is imperative for determining when to augment a slow tactic with a faster tactic. There may be instances when the system decides to utilize a faster, less effective tactic to augment a slower, but more effective tactic~\cite{moreno2015proactive, moreno2017adaptation}. For example, in a self-adaptive system some given tactic $a$ may have higher latency while producing a higher benefit than that of a faster tactic option $b$. The system may decide to implement both tactic options, knowing that the system could potentially realize the benefits of tactic $b$ while it is waiting for tactic $a$. When deciding to augment a slower tactic with a faster one, accounting for tactic latency volatility is essential to determine which tactic(s) should be used to augment a slower tactic.}
    


    \item \descStep{Ensuring the selection of the most appropriate tactic}{When selecting between multiple tactic options, ignoring tactic latency can be problematic. Consider a scenario where the system is deciding whether to use tactic $a$ or tactic $b$. If tactic $b$ is only slightly better than tactic $a$ in terms of instantaneous utility improvement, then the decision-making process would favor tactic $b$. However, if tactic $a$ is faster than tactic $b$, then tactic $a$ would start to accrue utility faster. This means that tactic $a$ may be the most optimal selection~\cite{moreno2015proactive}.}

\end{enumerate}

\subsection{Motivating Example} 
\label{sec: motivatingexample}

As a motivating example we will use a cloud-based self-adaptive system, based on a similar scenario defined by Moreno \etal\cite{moreno2017adaptation}. This example represents a multi-tier web application that is compromised a web server and database tier. The webserver(s) process a client's request and then access information stored in the database tier. To efficiently provide content while encountering variable workloads, the system can either add/remove servers from the pool or reduce optional content using the `dimmer' feature. This system has has a goal of maximizing utility while minimizing cost.

The SLA defines the target response time ($T$) and how utility ($U$) is calculated. The system incurs penalties if the target response time is not met and accrues rewards for meeting the target average response time against the measurement interval. 
The average response rate is $a$, the average response time is $r$, the maximum request is $k$ and the length of each interval is defined as $\tau$. Provided content is reduced as necessary using a dimmer value ($d$). Optional content has reward of ($R_O$), and produces a higher reward than mandatory content ($R_M$). We will slightly modify the equation to incorporate cost ($C$), which could be the monetary or energy cost:

\begin{equation} \label{eq: motivatingExample}
	U = \left\{ \begin{array}{rl}
 	(\tau a (dR_O+(1-d)R_M)/C~~~~r\leq T \\ 
  	(\tau \min (0,a-k)R_O)/C~~~~r> T 
       \end{array} \right.
\end{equation}





This system can account for increases in user traffic using two different tactics: (I) reducing the proportion of responses that include the optional content (dimmer), and (II) adding a new server. While reducing optional content will have negligible latency, adding a new server can take several minutes.

\vspace{0mm} \noindent \textbf{Tactic Latency Volatility} If the system anticipates that the response threshold will be surpassed in the immediate future, then the system could proactively start the tactic of adding a server to keep the response time under the defined threshold. Overestimating latency could result in scenarios where the system unnecessarily incurs additional cost, as servers would be `active' longer than necessary. Additionally, if the system determines that it is likely to surpass the defined response time threshold before a new server can be added, then the most appropriate system action may be to use the faster tactic that reduces the amount of optional content while it waits for a new server to be added. Improper tactic latency predictions can lead to situations where the system executes the tactic too soon or too late, or even selects the improper tactic for the encountered scenario. \emph{Accounting for tactic latency volatility is a paramount concern, especially when utilizing a proactive adaptation approach, or when utilizing complementary tactics.} 

\vspace{0mm} \noindent \textbf{Tactic Cost Volatility} In the provided scenario, it is important to account for cost volatility, especially since not accounting for cost volatility could lead to inaccurate utility calculations. In the event that cost is defined to be higher than what the system is actually encountering in the real-world, then this may lead to scenarios where optional ($O$) content is shown too infrequently. Conversely, if the cost is defined lower than what is being encountered in the real-world, then could lead to scenarios where optional ($O$) content is shown too frequently. \emph{A volatility aware solution that enables the system to more accurately predict cost would enable the system to make decisions that lead to more optimal outcomes.}







\vspace{0mm} \noindent \textbf{Reactive Specifications Monitoring} If the motivating example was a purely responsive system and did not employ any proactive functionality, it will only determine if the defined response threshold ($T$) is being surpassed at the current moment. This can be problematic since the tactic of adding additional resources to reduce response time in our example has latency, so the system will need to begin adapting \emph{before} the tactic is actually needed to be complete. Otherwise, the system will incur penalties or not realize rewards while it waits for the tactic to become available. \emph{A process that enables the system to better anticipate how it was going to act in accordance with the SLA would help the system to perform more optimally in dynamic environments.}

\section{Proposed TVA Process}
\label{sec:proposedTechniqueThesis}


Our TVA process consists of the following two phases: (I) Time-series forecasting with \emph{Autoregressive Integrated Moving Average} (ARIMA) and (II) Run-time model generation using Multiple Regression Analysis (MRA); where ARIMA predicts if system specifications in the SLA may be broken in the future and MRA creates run-time models for predicting tactic latency and cost.

\subsection{Autoregressive Integrated Moving Average}
The first component of the TVA process is ARIMA, which is a commonly used approach for the prediction of time series data. ARIMA is a generalization of the ARMA (autoregressive moving average) model, which accounts for non-stationary data using differencing. A full ARIMA treatment requires the following notation, \emph{ARIMA(p, d, q)}, where $p$ represents the number of lag observations included in the model, $d$ represents the degree of differencing, and $q$ represents the size, or \emph{order} of the moving average window.


This work uses a Box-Jenkins approach to find the best fit of a ARIMA model for predicting future time series values. This involves applying the following three steps:


\begin{enumerate}

\item \textbf{\textit{Identification: }}The first step is to determine the order of the ARIMA model, by utilizing \textit{differencing} to transform the potentially non-stationary data to stationary data. Mathematically, differences are shown as $y_{t}'=y_{t} - y_{t-1}$, where $y_t$ represents the current observation and $y_{t-1}$ represents the immediate-prior observation. Differencing allows the model to remove any changes in the levels of time series data, thus eliminating trend and seasonality. In some cases, differencing may need to be applied a second time to obtain stationary data. This second order differencing includes subtracting another term $y_t''=y_{t-1} - y_{t-2}$ to the ARIMA model. Our work used a differencing order of $d=1$, as there was a small amount of non-stationary time series data. Following this, classic autocorrelation and partial autocorrelation plots were used to determine the order of the autoregressive and moving average terms, which resulted in $p=1$ $q=0$, respectively.



\item \textbf{\textit{Estimation:} }In order to estimate parameters for the Box-Jenkins models, we must apply a solution that can numerically approximate nonlinear equations. As the goal was to minimize a loss or error term, we used the \textit{maximum likelihood estimation} (MLE) method over a nonlinear least squares estimation to determine the model's optimal parameters.


\item \textbf{\textit{Model Checking: }}The final step in applying a full treatment of the Box-Jenkins approach is to perform model checking. Since we have the ability to modify orders in the ARIMA model ($p$ and $q$) it is important that we optimize these parameters. In general, optimization should examine: I) if the model is overfit and II) residual errors. 
The former is crucial because it effects how generalizable our model is to other time series data, while the latter deals with how well the model performed in terms of predictions. After examining multiple ARIMA models by fine-tuning the three main parameters $p$, $d$ and $q$, the final ARIMA model for our work was $ARIMA(1,1,0)$. With this notation, our model is known as a \textit{differenced first-order autoregressive model}.



\end{enumerate}

\subsection{Multiple Regression Analysis}
The second component of our TVA approach deals with run time predictions of tactic latency and cost. As discussed previously, current self-adaptive processes consider these values to be static attributes. In real-world scenarios, this is an unlikely phenomena because certain events have different outcomes depending on the surrounding environment. 
For example, a UAV may need to determine the time it will take to transmit a critical file to its base station. This latency could drastically vary depending on the distance of the UAV to the base station, weather conditions and even component functionality. Therefore, the UAV must have a method of modeling its surrounding environment so it can accurately anticipate the length of time required to transmit the file.

In our work we applied this same ideology to predicting tactic latency and cost. There are many different types of regression models, and even machine-learning models that could be used to accomplish this. Initially, we examined a machine-learning approach called Bayesian Ridge Regression (BRR) which through a Bayesian process allows the model to train itself over time as more data becomes available. However, we found that the required feature space for BRR was too large and that the data we collected could not support it. Therefore, our TVA approach uses Multiple Regression Analysis (MRA) to support run time predictions of tactic latency and cost. Consider a classic regression model:

\begin{equation}
    y(x,w) = w_0 + \sum_{j=1}^{M-1} w_j x^j=w'x
\end{equation}

\noindent
where $x=(x^1,...,x^M)'$ with $x^0=1$ and $x^j$ being the $j$-th observation of the independent variable $x$. Given a set of $N$ training variables $(x_1,...,x_N)$ along with the observed responses $(t_1,...,t_N)$, we can solve for the best possible weights of this model through minimizing the error function:

\begin{equation}
E(w) = \frac{1}{2}\sum_{n=1}^{N}\{y(x_n,w)-t_n\}^2
\label{eq:mini}
\end{equation}

\noindent
$E(w)$ is a quadratic function of $w$ (the observation weight), which can be conveniently minimized, leading to:

\begin{equation}
	w^* = (X'X)^{-1}X't
\label{eq:weights}
\end{equation}

\noindent
where $X$ is the design matrix whose rows correspond to the observation vectors of the variables and $t=(t_1,...,t_N)'$ denotes the observed response values of $N$ variables. After minimizing Equation \ref{eq:mini}, we now have the estimated weights $w^*$ in Equation \ref{eq:weights}, which can be used by our regression model to predict tactic latency and cost.

It is important to note that regression analysis estimates the conditional expectation of the dependent variable, that is, the average value of the dependent variable when the predictor variables are fixed. Other related methods such as Necessary Condition Analysis (NCA) could be used to estimate the maximum value of the dependent variable, however, we decided that this was out of scope for our approach since estimating maximum values would be considered the \textit{worst-case scenario} for the tactic.

\vspace{1mm} \noindent \textbf{TVA Workflow}

TVA combines an ARIMA time series model with an MRA model to improve the self-adaptive process. In doing this, the system is able to properly maintain its defined specifications while also accounting for tactic volatility, leading to better adaptations and better overall performance. As shown in the pseudo code in Algorithm~\ref{alg:TVAWorkFlow}, the workflow of TVA is reasonably straightforward. The top level definitions under $procedure$ represent the specifications that are being monitored and any calculations that are made. For instance, $spec$ represents the current specification that we are analyzing, while $resp$ represents the response we obtain from performing the actual analysis. Furthermore, the $loop$ of the workflow represents the main component of TVA.

\begin{algorithm}
\caption{TVA Workflow}\label{alg:TVAWorkFlow}
\begin{algorithmic}[1]
\Procedure{Workflow}{}
\State $\textit{spec} \gets \text{the specification to analyze}$
\State $\textit{resp} \gets \text{quantified response from ARIMA analysis}$\\
loop:
\State $resp = analyzeSpecification(spec)$ \label{op:op0}
\If {$resp$ is potentiallyBroken}
\State $makeLatencyEstimate()$
\State $makeCostEstimate()$
\EndIf
\State \textbf{goto} loop
\EndProcedure
\end{algorithmic}
\end{algorithm}


We will next discuss a concrete example that examines the two primary steps of our TVA process as shown in the pseudo code. For this, we will continue to use the self-adaptive hosting service example described in Section \ref{sec: motivatingexample}.

\subsubsection*{\textbf{Step 1: Monitoring Specifications}}

The first step in our TVA approach is to gather and monitor the specifications defined in the SLA. Using the self-adaptive hosting service example, we will assume that it has been configured to record the response time every six seconds. We will also assume that the system's SLA specification defines that a penalty will be incurred if response times do not stay under a 0.7 second threshold. Unfortunately, the majority of self-adaptive processes act in a reactive manner, so the system would not be able to adapt until after the threshold was surpassed. 
Through the application of the time series model ARIMA discussed in this section, we will enable the system to act more proactively by allowing it to anticipate future specification values.

Given the specification and parameters required to make decisions, historical data is used to determine parameters to the ARIMA models (Algorithm~\ref{alg:TVAWorkFlow}, Line~\ref{op:op0}), which provide anticipated future values. For example, in the hosting service when the response time specification is analyzed with ARIMA, the ``response'' is the forecast value. If this forecast value determines that the specification may be broken in the future or is close to being broken, we continue with the rest of our TVA approach.

\subsubsection*{\textbf{Step 2: Tactic Latency and Cost Prediction}}

When the system has determined that it needs to adapt, whether it be because a specification \textit{will} or \textit{may} be broken, our approach then uses regression analysis to predict tactic latency and tactic cost. For example, if our ARIMA analysis from step 1 determines that the hosting service needs to adapt, our TVA approach then predicts tactic latency and cost for all available adaptation tactics. This could entail predicting the latency and cost for a tactic that adds extra computing resources so that response time can be reduced. These predictions are made through a Multiple Regression Analysis (MRA) model. A benefit of focusing on predicting tactic latency and tactic cost is that this can be easily adopted into many existing decision-making processes. For example, one simple option is for the predicted values to merely replace the static, predefined values in the system's decision-making process.

\section{VALET: VolAtiLity EmulaTor Tool}
\label{sec: VALETTool}

An additional contribution of this work is the development of the VolAtiLity EmulaTor (\emph{VALET}) tool to generate real-world tactic volatility data to enable the evaluation of our TVA process. While existing datasets such as `The Internet Traffic Archive'\cite{TheInternetTrafficArchive_URL} are commonly used when evaluating self-adaptive processes~\cite{gandhi2014adaptive, dhanapal2017effective}, these are limited as they do not contain an adequate amount of tactic volatility necessary to evaluate our work. To create a sufficient dataset to demonstrate the benefits of TVA, VALET provides tactic volatility data in the form of latency and cost. The tool and generated dataset is available on the project website: \ToolURL

The first action performed in VALET utilizes two physical Raspberry Pi devices, with one acting as the operator and the other as a monitoring instrument. The operator gathers latency data by recording the time required to download an identical 75MB file at one minute intervals, from three different download mirrors around the world. To determine energy usage for this operation, the monitoring instrument collects and records its own energy usage as well as that of the operator. Then, the monitoring Pi calculates the difference between the operating and monitoring Pi's power before and after each download as the overall energy used for the operation. This action provides both real-world latency and cost information. Real-world volatility is also introduced by variations in network speeds that impact the time required to download the file.

The second action performed by VALET obtains an additional form of tactic latency and cost values by performing a `grep' activity. After the file is downloaded, a simple \emph{grep} command searches files contents for a specific string. The amount of time and energy required to perform this task is recorded. 

\begin{figure}[h]
	\centering
     \includegraphics[width=0.9\linewidth]{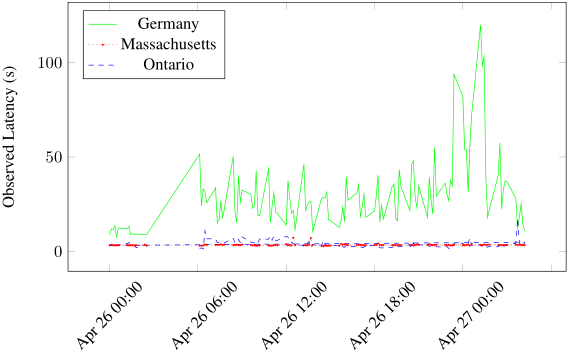}
    \caption{9-Hour+ Snippet of Latency Data}
    \label{fig:timeseries1}
\end{figure}

These operations are performed at one minute intervals over the course of an entire day, creating over 1,400 records of tactic volatility data daily. 
Figure~\ref{fig:timeseries1} represents a portion of the time series data gathered from our operations, and also visualizes the volatile latency values that we obtained. As shown, the latency times gathered from Germany were quite volatile, with some spikes in volatility in Massachusetts and Ontario download times. 

VALET benefits the software engineering community by enabling developers and researchers to perform evaluations using real-world, time-series data containing tactic volatility. Although VALET represents a specific example of a simple self-adaptive system, the created data is generic enough that it can be used to conduct preliminary evaluations of other machine-learning processes and tactic volatility aware self-adaptive processes.




\section{Evaluation}
\label{sec: Evaluation}

\newcommand{\RQA}{How does not accounting for tactic volatility affect the decision-making process?} 
\newcommand{\RQB}{How effective is ARIMA in allowing the system to monitor system specifications?} 
\newcommand{\RQC}{How effective is MRA in predicting tactic latency and cost at run time?} 
\newcommand{\RQD}{Does using TVA provide substantial improvement to the self-adaptive process over simply using static values for latency and cost?} 

This evaluation addresses the following research questions:

\hangindent=2.5em 
\textbf{RQ1.} \emph{\RQA} Using the statistical tool R, we demonstrate the negative impact of not accounting for tactic latency volatility and tactic cost volatility in the self-adaptive process.

\hangindent=2.73em 
\textbf{RQ2.} \emph{\RQB} In our experiments using time series analysis, we demonstrate that time series forecasting with ARIMA helps the system to become more proactive in maintaining specifications defined in the SLA.

\hangindent=2.62em 
\textbf{RQ3.} \emph{\RQC} In evaluating our multiple regression model, we demonstrate that strong predictive power exists throughout our experiments when estimating tactic latency and cost, even when faced with varying amounts of volatility in the gathered data.

\hangindent=2.47em 
\textbf{RQ4.} \emph{\RQD} In comparing our approach against existing self-adaptive approaches, such as those that consider tactics to have static latency and cost values, we demonstrate the substantial improvement to the decision-making process that our TVA process provides.

\subsection{\textbf{Experimental Data Analysis}}

To analyze the results found from our VALET experiments, we used the statistical metrics of Root Mean Square Error (RMSE) and Mean Absolute Error (MAE) to evaluate the systematic benefits of our TVA approach. Where $y_i$ is the predicted value and $t_i$ is the actual value at time $i$, RMSE and MAE are defined as:

\begin{equation} 
  		RMSE = [ \sum\limits_{i=1}^{N} (y_i - t_i)^2 / N]^{1/2}
  		\label{eq: rmse}
\end{equation}

\begin{equation} 
  		MAE = \dfrac{\sum\limits_{i=1}^{n} |y_i - t_i|}{n}
  		\label{eq: mae}
\end{equation}

These metrics allowed us to determine TVA's ability to reduce uncertainty and increase the effectiveness of the decision-making process by examining the prediction accuracy of our time series and machine learning models. RMSE provides a better search landscape for determining model parameters and was used to determine how well our models performed in terms of predictive ability, as larger prediction errors become more pronounced due to the squaring of such errors. On the other hand, MAE was used to examine the absolute value of these error differences. MAE was also used for looking at forecasting errors in time series analysis, which is one of its most common uses \cite{mae}.

\subsection{\textbf{Results}}

\noindent \textbf{RQ1: \RQA} We first explored RQ1 by performing a proof of concept evaluation using the statistical tool R, where we emulated the negative effects of poor decision-making by a self-adaptive system. We began by defining two tactics (A $\&$ B) that had arbitrarily defined costs associated with their latency times, where if the tactic took $X$ amount of time to execute, it had $C$ units of cost.

\begin{table}[h]
\begin{center}
\caption{Characteristics of Sample Tactics for Proof of Concept Simulation}
\vspace{2mm}
\label{table: impact}

  \begin{tabular}{ c | c | c | c | c } \hline
 \toprule
  \bfseries & \bfseries Cost & \bfseries Distribution & \bfseries Average & \bfseries SD \\  \midrule 
    \bfseries Tactic A & 5 & Positively Skewed & 3 & 0.5 \\ \hline
    \bfseries Tactic B & 7 & Normal & 3 & 0.5 \\ 
\bottomrule
   \end{tabular}
  \end{center}
\end{table}
\vspace{-1mm}

We next generated a normal distribution and positively skewed distribution with the same mean and standard deviation to represent latency times. We gave the tactic with the higher cost (B) the normal distribution, while the tactic with the lower cost (A) was given the positively skewed distribution. By using two divergent values, we are able to demonstrate the impact of not accounting for tactic volatility in a near-perfect environment (the normal distribution) and in a volatile environment (the positively skewed distribution). It is possible that both latency and cost do not follow these types of distributions, but because they are two extremes, it was sufficient for our proof of concept. Table~\ref{table: impact} shows the characteristics of each tactic.

\definecolor{blue-violet}{rgb}{0.54, 0.17, 0.89}
\definecolor{candypink}{rgb}{0.89, 0.44, 0.48}
\definecolor{ceil}{rgb}{0.57, 0.63, 0.81}
\definecolor{fuzzywuzzy}{rgb}{0.8, 0.4, 0.4}

\begin{figure}[h!]
	\centering
	\resizebox{.9\linewidth}{.65\linewidth}{
     
\begin{tikzpicture}
\begin{axis}[
  ylabel=Frequency of Cost,
  legend style={font=\fontsize{8.5}{8.5}\selectfont},
  xlabel=Cost of executing a tactic, 
  legend image code/.code={
        \draw [#1] (0cm,-0.1cm) rectangle (0.6cm,0.1cm);
    },
]

 \addplot[ybar,bar width=11.5pt,fill=none,opacity=0.8, pattern=horizontal lines] table[col sep=comma,x index=0,y index=1] {data1.txt};\addlegendentry{Tactic A}

 \addplot[ybar,bar width=11.5pt,fill=ceil,opacity=0.6] 
  table[col sep=comma,x index=0,y index=2] {data1.txt};\addlegendentry{Tactic B}

\end{axis}
\end{tikzpicture}
} 
    \caption{Overall costs of executing a tactic in the R simulations.}
    \label{fig:needForAccountingForVolatility}
\end{figure}
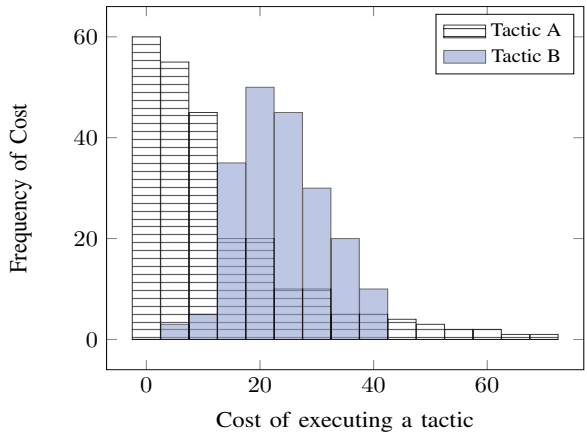

After generating the characteristics of each tactic, we then performed 100 simulations of random sampling from each tactic's latency and multiplied the sampled latency value by its associated cost. This enabled us to build a simulated distribution of ``tactic executions'', where the same tactic is executed every time, but with different latency values. This also allowed us to gather overall cost values that could be used to compare how varying data distributions can impact the predictability of cost for executing a tactic without accounting for any form of volatility.

In our analysis, we found that tactic $B$ may not have had the lowest cost values in the simulations, but it was significantly more consistent (Figure \ref{fig:needForAccountingForVolatility}). However, tactic $B$'s data followed a normal distribution, which is an unlikely occurrence for many real-world systems. Furthermore, even though tactic $A$ had a lower cost of execution with a value of five, it resulted in much more sporadic, and extremely high, overall costs to the system.

\noindent \textbf{Outcome:} \textit{Our proof of concept simulations in R successfully demonstrate that accounting for tactic volatility is essential in self-adaptive systems, especially when the system is known to have unpredictable behavior.}

\noindent \textbf{RQ2: \RQB} To evaluate our ARIMA model for time-series forecasting, we used the energy data collected from our VALET tool. Although we had been using a self-adaptive hosting service as an example in this paper, specifically one that monitored response time, the energy usage data is of the same concept. For both, data is gathered over a period of time at equal intervals, thus qualifying it as time-series data. However, the only difference with this analysis is that we did not consider the energy usage when VALET is downloading the file. Therefore, the data used in this research question is strictly the energy usage fluctuations when the device was idle and \textit{not} performing a tactic activity.

To specifically address this research question, we had to loop through the data multiple times to ensure that ARIMA could provide sustainable time-series forecasting predictions. We did this by creating 50 experiments using a randomly selected 90\% portion of our data as training data, and the other 10\% as test data. This type of process can also be seen as a form of \textit{k-cross validation}, which ensures that each data point is included in the training set at least once. After performing this validation, we then calculated both RMSE and MAE values to determine the predictive ability of the ARIMA model.

\begin{figure}[h]
    \centering
	\resizebox{.9\linewidth}{.65\linewidth}{
    
\begin{tikzpicture}
\begin{axis}[
ymin=-0.1, ymax=0.25,
xtick pos=left,
ytick pos=left,
ylabel=MAE Value,
xlabel=Simulation \textit{N}
]
	\addplot[only marks, blue] coordinates {
	(0,0.0807146) (1,0.0713026) (2,0.05549) (3,0.146) (4,0.07891) (5,0.1082) (6,0.08767) (7,0.05549) (8,0.1035863) (9,0.090701) (10,0.15396) (11,0.118238) (12,0.103863) (13,0.047307) (14,0.05549) (15,0.123976) (16,0.0943001) (17,0.080716) (18,0.0709601) (19,0.077146) (20,0.06607) (21,0.0713076) (22,0.0803746) (23,0.06401) (24,0.07282) (25,0.06978) (26,0.079281) (27,0.0637641) (28,0.06958) (29,0.11878) (30,0.117531) (31,0.0667258) (32,0.06926) (34,0.06696) (35,0.1534396) (36,0.0676496) (37,0.1181738) (38,0.07146) (39,0.027546) (40,0.1196) (41,0.0943601) (42,0.06031) (43,0.060376) (44,0.15896) (45,0.055849) (46,0.078282) (47,0.11101) (48,0.075258) (49,0.07281)};\addlegendentry{HMM}
	
	\addplot[only marks, mark=o,draw=black, red] coordinates {
	(0,0.05629) (1,0.04103) (2,0.04332) (3,0.1046) (4,0.06321) (5,0.07871) (6,0.066720) (7,0.04412) (8,0.06899) (9,0.03029) (10,0.0987) (11,0.07152) (12,0.06113) (13,0.02363) (14,0.03243) (15,0.07598) (16,0.065563) (17,0.05397) (18,0.03456) (19,0.05563) (20,0.0602) (21,0.04332) (22,0.075563) (23,0.05661) (24,0.044503) (25,0.044879) (26,0.05234) (27,0.033289) (28,0.02236) (29,0.09956) (30,0.09112) (31,0.04631) (32,0.02223) (34,0.0510) (35,0.101123) (36,0.04663) (37,0.09445) (38,0.04976) (39,0.0156) (40,0.10112) (41,0.051238) (42,0.04361) (43,0.031126) (44,0.1098) (45,0.04629) (46,0.066731) (47,0.07134) (48,0.059713) (49,0.03164)};\addlegendentry{ARIMA}
    
    \draw[ultra thick, black] (axis cs:\pgfkeysvalueof{/pgfplots/xmin},0) -- (axis cs:\pgfkeysvalueof{/pgfplots/xmax},0);
    
\end{axis}
\end{tikzpicture}

 } 

\caption{MAE Values Over 50 Experiments Using ARIMA vs HMM for Time Series Forecasting}
\label{fig:maeHMM}

\end{figure}
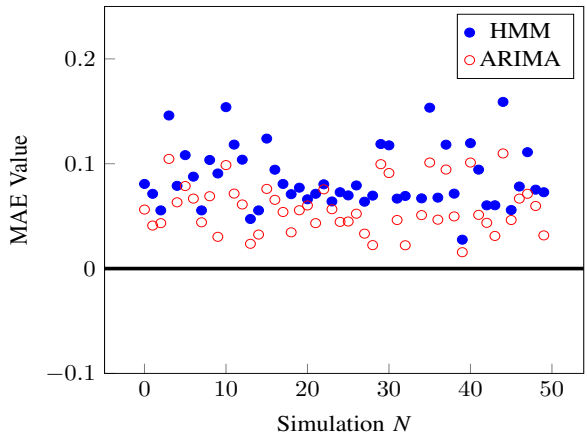

Figure \ref{fig:maeHMM} shows the MAE results from our ARIMA model compared to a previously used Hidden Markov Model (HMM). Each simulation was independently performed, so no patterns should be inferred from the left to right sequence. Over the course of the experiments we saw fairly stable MAE values, represented by the small range between the lowest MAE value and the highest. Furthermore, the ARIMA model was not only successful by itself, but also better than that of the HMM model. We believe this occurred as we did not have enough features in our model to allow for a full treatment of a HMM.


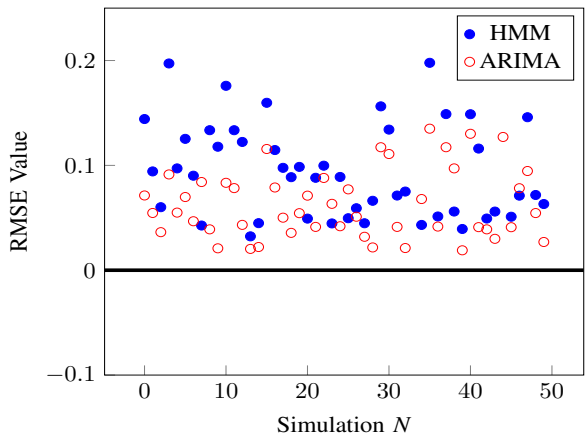
\begin{figure}[h]
    \centering
	\resizebox{.9\linewidth}{.65\linewidth}{
    
\begin{tikzpicture}
\begin{axis}[
ymin=-0.1, ymax=0.25,
xtick pos=left,
ytick pos=left,
ylabel=RMSE Value,
xlabel=Simulation \textit{N}
]
	\addplot[only marks, blue] coordinates {
	(0,0.14420) (1,0.09424) (2,0.06012) (3,0.1972) (4,0.09712) (5,0.1253) (6,0.09015) (7,0.04263) (8,0.13356) (9,0.1178) (10,0.1759) (11,0.13354) (12,0.12236) (13,0.03231) (14,0.04489) (15,0.1597) (16,0.11463) (17,0.09765) (18,0.088746) (19,0.09856) (20,0.04923) (21,0.088163) (22,0.09971) (23,0.04463) (24,0.088963) (25,0.0496) (26,0.05913) (27,0.04480) (28,0.06623) (29,0.15630) (30,0.13412) (31,0.07123) (32,0.07503) (34,0.04320) (35,0.19783) (36,0.051261) (37,0.148913) (38,0.05597) (39,0.039336) (40,0.14879) (41,0.11603) (42,0.04931) (43,0.05594) (44,0.18936) (45,0.051023) (46,0.071136) (47,0.14593) (48,0.07176) (49,0.06314)};\addlegendentry{HMM}
	
	\addplot[only marks, mark=o,draw=black, red] coordinates {
	(0,0.07130) (1,0.0546) (2,0.03630) (3,0.09136) (4,0.05496) (5,0.06981) (6,0.046720) (7,0.08412) (8,0.03899) (9,0.0209) (10,0.08337) (11,0.07832) (12,0.04326) (13,0.02033) (14,0.02213) (15,0.11563) (16,0.07891) (17,0.05012) (18,0.03563) (19,0.054413) (20,0.07113) (21,0.04132) (22,0.08801) (23,0.06332) (24,0.04195) (25,0.07713) (26,0.05103) (27,0.03190) (28,0.02173) (29,0.1173) (30,0.11083) (31,0.04133) (32,0.02116) (34,0.0679) (35,0.1350) (36,0.04165) (37,0.11732) (38,0.09713) (39,0.0190) (40,0.13013) (41,0.04109) (42,0.0389903) (43,0.029948) (44,0.1271) (45,0.04103) (46,0.07821) (47,0.09463) (48,0.054522) (49,0.02693)};\addlegendentry{ARIMA}
    
    \draw[ultra thick, black] (axis cs:\pgfkeysvalueof{/pgfplots/xmin},0) -- (axis cs:\pgfkeysvalueof{/pgfplots/xmax},0);
    
\end{axis}
\end{tikzpicture}

}

\caption{RMSE values over 50 experiments using ARIMA vs HMM for Time Series Forecasting}
\label{fig:rmseHmmArima}

\end{figure}

As discussed previously, larger prediction errors will become more pronounced and smaller prediction errors will become less pronounced when using RMSE. Due to this, we expected the RMSE graph to be slightly more dispersed compared to MAE. In examining the RMSE differences between HMM and ARIMA, we found exactly this. Looking at Figure \ref{fig:rmseHmmArima} closely, we can see how the RMSE values have more variation than the MAE values in Figure \ref{fig:maeHMM}. However, this does not mean there was less predictive power or that the model is less useful. We report on both to show the differences in possible inferences. For example, in applying our TVA technique the system's engineers may care more about larger prediction errors; deeming RMSE a more powerful statistic for determining predictive ability of their models. Conversely, if the system's engineers want more of a ``man-in-the-middle'' statistic, they may deem MAE more appropriate. For our experiments, using RMSE or MAE would lead to the same conclusion -- both statistics clearly favor the ARIMA model over the HMM model for time series forecasting.

\noindent \textbf{Outcome:} \textit{Through an evaluation process using time series data, TVA demonstrated it's ability to positively support proactive adaptations.}\\


\noindent \textbf{RQ3: \RQC} Unpredictability is considered to be an undesirable trait of a self-adaptive system and is frequently associated with much of the uncertainty that surrounds self-adaptive systems~\cite{rodrigues2018learning, tomforde2016organic, Kinneer2018ManagingUI, esfahani2014management}. To determine how well TVA can improve the predictability of a self-adaptive system, we examined the prediction errors across our tactic latency data and our tactic cost data. For space considerations and the use of RMSE being more appropriate in this context, this research question does not report MAE values.


Unlike RQ \#2 where we only examined the energy usage for when the Raspberry Pi was idle, RQ \#3 only examined the energy usage when the device \textit{was} performing the file download. If we had used the energy usage data from when the device was idle, our models would have been invalid. This is because the file download represented a tactic of gathering more information, thus any energy data gathered while the device was idle did not represent tactic latency.

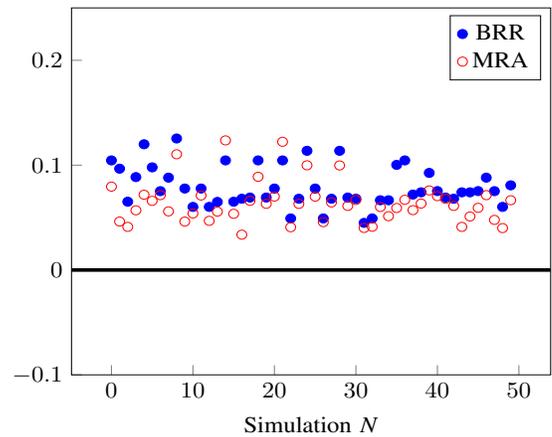
\begin{figure}[h]
    \centering
 	\resizebox{.9\linewidth}{.65\linewidth}{
 	
\begin{tikzpicture}
\begin{axis}[
ymin=-0.1, ymax=0.25,
xtick pos=left,
ytick pos=left,
ylabel=RMSE Value,
xlabel=Simulation \textit{N}
]
	\addplot[only marks, blue] coordinates {
	(0,0.104526215611) (1,0.09663001) (2,0.0651396100502) (3,0.088610036) (4,0.1199093) (5,0.0979030) (6,0.0752027817414) (7,0.088082200573) (8,0.125437336115) (9,0.0777104) (10,0.0601853376522) (11,0.0777109489224) (12,0.0601853376522) (13,0.0649311136285) (14,0.104526215611) (15,0.0651396100502) (16,0.067980046989) (17,0.0691105310622) (18,0.104526215611) (19,0.0691105310622) (20,0.0777109489224) (21,0.104526215611) (22,0.0491833826094) (23,0.067980046989) (24,0.113685402858) (25,0.0777109489224) (26,0.0491833826094) (27,0.067980046989) (28,0.113685402858) (29,0.0691105310622) (30,0.067980046989) (31,0.0449456839679) (32,0.0491833826094) (33,0.066531787513) (34,0.066531787513) (35,0.100268434489) (36,0.104526215611) (37,0.0719623111926) (38,0.0740933208235) (39,0.0925966318751) (40,0.0752655663011) (41,0.0691105310622) (42,0.067980046989) (43,0.0740933208235) (44,0.0740933208235) (45,0.0752655663011) (46,0.088082200573) (47,0.0752027817414) (48,0.0601853376522) (49,0.0807571593297)
	};\addlegendentry{BRR}
    
    \addplot[only marks, mark=o,draw=black, red] coordinates {
    (0,0.07953) (1,0.04623) (2,0.04113) (3,0.05693) (4,0.0717930) (5,0.06593) (6,0.071394) (7,0.055930) (8,0.11036) (9,0.046179) (10,0.053910) (11,0.071063) (12,0.0469003) (13,0.0556971) (14,0.123609) (15,0.053608) (16,0.033691) (17,0.065960) (18,0.088913) (19,0.063120) (20,0.0700356) (21,0.12230) (22,0.0407993) (23,0.063061) (24,0.099861) (25,0.0699789) (26,0.045630) (27,0.064430) (28,0.099730) (29,0.061130) (30,0.0667933) (31,0.0401136) (32,0.0412369) (33,0.0601137) (34,0.051306) (35,0.05913) (36,0.066971) (37,0.05710310) (38,0.0633719) (39,0.075906) (40,0.0704462) (41,0.0679003) (42,0.06113086) (43,0.0411036) (44,0.05103791) (45,0.05930036) (46,0.071393) (47,0.0477913) (48,0.0399713) (49,0.0665130)
	};\addlegendentry{MRA}
    
    \draw[ultra thick, black] (axis cs:\pgfkeysvalueof{/pgfplots/xmin},0) -- (axis cs:\pgfkeysvalueof{/pgfplots/xmax},0);
    
\end{axis}
\end{tikzpicture}

}

\caption{RMSE Values for Tactic Cost Predictions of MRA vs BRR}
\label{fig:rmseCostValet}

\end{figure}



Figure \ref{fig:rmseCostValet} demonstrates why MRA is more appropriate for TVA. Over the course of the 50 simulations, MRA reported lower RMSE values than BRR in almost every case. In cases where BRR did outperform MRA, the differences were fairly negligible. However, the plots are closer together than what we would have initially expected. We believe this was caused by only having a few cases of extreme volatility in the cost data, therefore not allowing the two models to really differentiate themselves.  

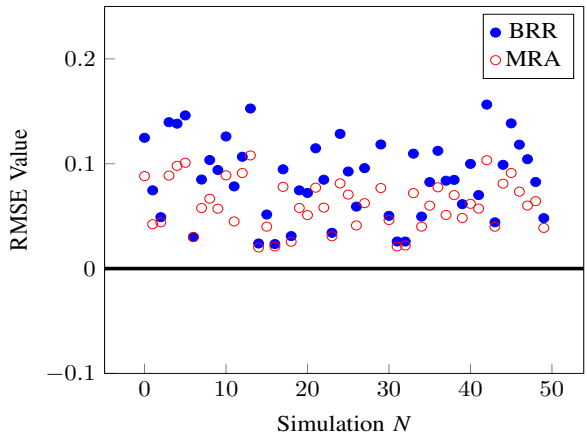
\begin{figure}[h]
    \centering
	\resizebox{.9\linewidth}{.65\linewidth}{
    
\begin{tikzpicture}
\begin{axis}[
ymin=-0.1, ymax=0.25,
xtick pos=left,
ytick pos=left,
ylabel=RMSE Value,
xlabel=Simulation \textit{N}
]
	\addplot[only marks, blue] coordinates {
	(0,0.124746) (1,0.0746) (2,0.049) (3,0.1396) (4,0.1382) (5,0.146082) (6,0.03001267) (7,0.0849) (8,0.103511827663) (9,0.09401) (10,0.126) (11,0.07838) (12,0.10663) (13,0.15267) (14,0.0239) (15,0.05156) (16,0.0233701) (17,0.09467) (18,0.031) (19,0.0746) (20,0.0721) (21,0.114746) (22,0.084746) (23,0.0341) (24,0.1285) (25,0.09258) (26,0.0591) (27,0.0958) (29,0.11838) (30,0.05031) (31,0.0258) (32,0.0258) (33,0.1096) (34,0.0496) (35,0.082496) (36,0.112296) (37,0.0838) (38,0.0846) (39,0.06146) (40,0.09986) (41,0.0701) (42,0.15641) (43,0.0441) (44,0.09896) (45,0.13849) (46,0.1182) (47,0.10431) (48,0.08258) (49,0.0481)};\addlegendentry{BRR}
    
    \addplot[only marks, mark=o,draw=black, red] coordinates {
    (0,0.088193) (1,0.0423360) (2,0.044160) (3,0.088693) (4,0.09799) (5,0.100893) (6,0.030113) (7,0.057719) (8,0.06663) (9,0.0571411) (10,0.08893) (11,0.044976) (12,0.091130) (13,0.107910) (14,0.020119) (15,0.039970) (16,0.021126) (17,0.077913) (18,0.0255617) (19,0.0577963) (20,0.0510030) (21,0.0771682) (22,0.05813645) (23,0.0307947) (24,0.0812409) (25,0.0705581) (26,0.041130) (27,0.062405) (29,0.076810) (30,0.04630) (31,0.021130) (32,0.022003) (33,0.071993) (34,0.0400191) (35,0.06003215) (36,0.077506) (37,0.0511360) (38,0.0700363) (39,0.04822613) (40,0.0617752) (41,0.05722036) (42,0.1033608) (43,0.0399613) (44,0.08100231) (45,0.0911963) (46,0.07339192) (47,0.0601123) (48,0.0643520) (49,0.03881095)
	};\addlegendentry{MRA}
    
    \draw[ultra thick, black] (axis cs:\pgfkeysvalueof{/pgfplots/xmin},0) -- (axis cs:\pgfkeysvalueof{/pgfplots/xmax},0);
    
\end{axis}
\end{tikzpicture}

}

\caption{RMSE values for tactic latency predictions of MRA vs BRR}
\label{fig:rmseLatencyValet}

\end{figure}

As shown in Figure \ref{fig:rmseLatencyValet}, we observed very similar results between the RMSE values calculated for tactic latency predictions and those calculated for tactic cost. In most cases, prediction errors with MRA were much smaller than those of BRR, and in examining the figures closely, the differences in volatility that were experienced can be observed. As mentioned previously, there were not as many extreme cases of volatility in the tactic cost data. However, within the tactic latency data we saw many cases of volatility, with some cases being extreme. This can partially be seen in Figure \ref{fig:rmseLatencyValet} since more volatility will likely lead to larger prediction errors. Thus, it came at no surprise that the RMSE values for tactic latency had a wider-spread than those associated with tactic cost. 

In order to say the MRA model can be generalized to other data sets for predicting tactic latency and cost, observing consistent RMSE values regardless of the data it was modeling was imperative. Figure \ref{fig:rmseCostValet} and Figure \ref{fig:rmseLatencyValet} demonstrate that not only were RMSE values low for both data sets, but the values were stable. Therefore, MRA was able to handle the volatile latency data and cost data collected for these experiments when making its predictions. 

\noindent \textbf{Outcome:} \textit{The RMSE results gathered from addressing this research question demonstrate that MRA is able to handle tactic volatility when predicting tactic latency and cost. Throughout our experiments, MRA consistently provided stable predictive power, even in the presence of volatile data.}

\noindent \textbf{RQ4: \RQD} Thus far, we have demonstrated the benefits of using an ARIMA time series model and a MRA model in a self-adaptive process to predict tactic latency and cost, while also maintaining specifications defined in the SLA. However, to provide further confidence in our TVA approach, we also compared it to current self-adaptive processes. Since current processes consider tactic latency and cost to be static values, there is no one specific work that TVA can be compared to. Rather, for this research question, we compared the results from our approach to what we consider the ``baseline'' model defined below:

\begin{framed}
\textit{\textbf{Baseline Model}: A model that uses a simple average of previous tactic latency and cost values as its prediction process for estimating future tactic behavior.}
\end{framed}

While utilizing the average value for latency and cost values may seem like a naive comparison, in many cases it can be a be a strong predictor especially when compared a static value (as in other current self-adaptive processes) on a dynamic time series. Similar to RQ \#2, we will also report the MAE and RMSE values for model comparison.



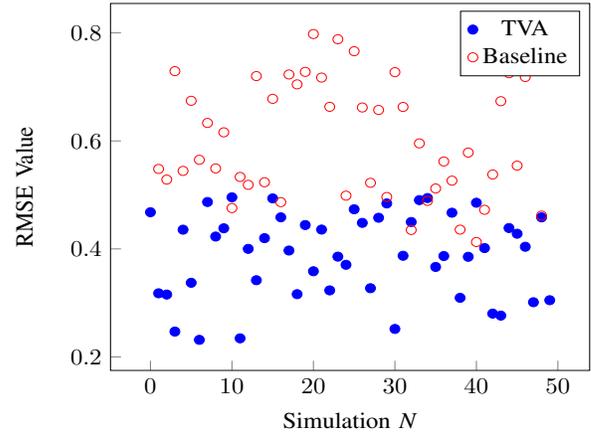
\begin{figure}[h]
    \centering
	\resizebox{.9\linewidth}{.65\linewidth}{	
    
\begin{tikzpicture}
\begin{axis}[
xtick pos=left,
ytick pos=left,
ylabel=RMSE Value,
xlabel=Simulation \textit{N}
]
	\addplot[only marks, blue] coordinates {
	(0,0.4679022) (1,0.3178) (2,0.3155) (3,0.2469) (4,0.4356) (5,0.33728) (6,0.2316) (7,0.48687) (8,0.42289) (9,0.4382) (10,0.4956) (11,0.2344) (12,0.4001) (13,0.3420) (14,0.4199) (15,0.49354) (16,0.4585) (17,0.3970) (18,0.3162) (19,0.4442) (20,0.3586) (21,0.43581) (22,0.3232) (23,0.3857) (24,0.3706) (25,0.4734) (26,0.4482) (27,0.32721) (28,0.4577) (29,0.48418) (30,0.25178) (31,0.387339) (32,0.4498) (33,0.4904) (34,0.4942) (35,0.3666) (36,0.38681) (37,0.46701) (38,0.30941) (39,0.3854) (40,0.4855) (41,0.40154) (42,0.2802) (43,0.27649) (44,0.43867) (45,0.42811) (46,0.4040) (47,0.30126) (48,0.45843) (49,0.30499) };\addlegendentry{TVA}
    
    \addplot[only marks, mark=o,draw=black, red] coordinates {
    (0.6066) (1,0.5481) (2,0.52841) (3,0.72919) (4,0.5445) (5,0.6743) (6,0.56503) (7,0.63325) (8,0.54903) (9,0.61590) (10,0.4756) (11,0.5333) (12,0.51868) (13,0.72003) (14,0.5234) (15,0.67793) (16,0.4866) (17,0.72304) (18,0.704710) (19,0.72784) (20,0.79789) (21,0.7173) (22,0.66314) (23,0.78829) (24,0.49865) (25,0.76610) (26,0.66190) (27,0.52240) (28,0.6573) (29,0.49610) (30,0.72732) (31,0.66288) (32,0.4350913) (33,0.59541) (34,0.48901) (35,0.51190) (36,0.56188) (37,0.52642) (38,0.43569) (39,0.578414) (40,0.41275) (41,0.47245) (42,0.5378) (43,0.673521) (44,0.72538) (45,0.55426) (46,0.7183510) (47,0.780043) (48,0.461672) (49,0.79306)
    };\addlegendentry{Baseline}

\end{axis}
\end{tikzpicture}

}

\caption{Baseline Value Approach vs TVA - RMSE Differences}
\label{fig:rmseComparisons}

\end{figure}

As shown in Figure \ref{fig:rmseComparisons}, TVA was able to obtain substantially lower RMSE values. In comparison, TVA had an average RMSE value of 0.0396 while the baseline approach had an average RMSE value of 0.0694. Also represented in this diagram is the ability of TVA to handle volatility. While the spread of RMSE values remained fairly consistent for TVA, the baseline approach saw a much wider spread; a direct result of only using a static value for latency and cost. Although there were cases where the baseline approach did outperform TVA (7 in total of 50 experiments), the differences in these values were fairly negligible. 

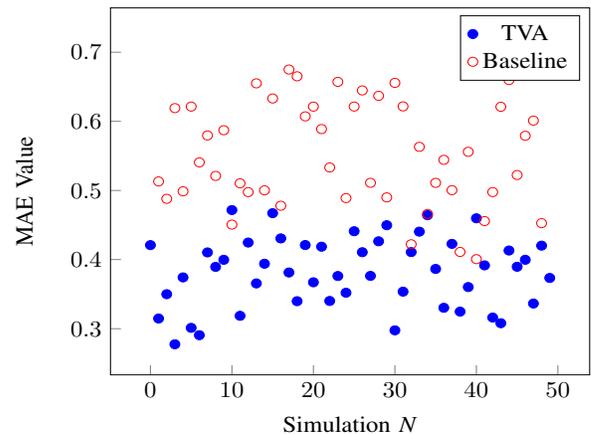
\begin{figure}[h]
    \centering
	\resizebox{.9\linewidth}{.65\linewidth}{
    
\begin{tikzpicture}
\begin{axis}[
xtick pos=left,
ytick pos=left,
ylabel=MAE Value,
xlabel=Simulation \textit{N}
]
	\addplot[only marks, blue] coordinates {
	(0,0.42090) (1,0.31469) (2,0.34992) (3,0.27745) (4,0.3742) (5,0.30119) (6,0.2906) (7,0.41035) (8,0.38942) (9,0.399682) (10,0.47163) (11,0.318643) (12,0.42463) (13,0.3655) (14,0.3940) (15,0.46712) (16,0.4306) (17,0.38132) (18,0.33978) (19,0.42103) (20,0.36713) (21,0.41863) (22,0.34012) (23,0.37620) (24,0.35203) (25,0.441003) (26,0.41064) (27,0.376403) (28,0.42643) (29,0.44970) (30,0.29761) (31,0.35360) (32,0.410793) (33,0.44039) (34,0.4642) (35,0.386460) (36,0.33025) (37,0.42273) (38,0.324692) (39,0.36024) (40,0.45973) (41,0.391630) (42,0.316098) (43,0.307913) (44,0.41309) (45,0.38961) (46,0.39972) (47,0.336412) (48,0.42016) (49,0.37340) };\addlegendentry{TVA}
    
    \addplot[only marks, mark=o,draw=black, red] coordinates {
    (0.55430) (1,0.51306) (2,0.48790) (3,0.61913) (4,0.49873) (5,0.62136) (6,0.54063) (7,0.57959) (8,0.521033) (9,0.58716) (10,0.4506) (11,0.51033) (12,0.49760) (13,0.65493) (14,0.500178) (15,0.63306) (16,0.47790) (17,0.67492) (18,0.66513) (19,0.60719) (20,0.621303) (21,0.5888130) (22,0.533261) (23,0.65709) (24,0.48893) (25,0.621396) (26,0.64469) (27,0.51103) (28,0.63688) (29,0.48996) (30,0.655643) (31,0.62163) (32,0.422130) (33,0.56310) (34,0.46604) (35,0.510996) (36,0.544306) (37,0.500397) (38,0.411036) (39,0.555973) (40,0.400613) (41,0.4557136) (42,0.4976310) (43,0.62106) (44,0.65971) (45,0.52230) (46,0.579162) (47,0.6010396) (48,0.45269) (49,0.7198703)
    };\addlegendentry{Baseline}

\end{axis}
\end{tikzpicture}

}

\caption{Baseline Value Approach vs TVA - MAE Differences}
\label{fig:maeComparisons}

\end{figure}

In examining the MAE values shown in Figure \ref{fig:maeComparisons}, we found that MAE did not behave as we expected. In RQ \#2, it is fairly evident that RMSE made larger prediction errors larger, and smaller prediction errors smaller. However, in this research question, those kinds of results are not as evident. In examining the MAE values from TVA, we can see the plot is a bit more condensed than then RMSE plot, but the MAE values for the baseline model did not follow the same trend as strongly. We believe this occurred because the baseline model's prediction method was extremely poor, justified by both the RMSE and MAE plots. Regardless, TVA was still superior in predictive ability compared to the baseline model.

\vspace{0mm}
\noindent \textbf{Outcome:} \textit{These findings demonstrate that TVA is beneficial for decision-making processes in self-adaptive systems, and offer a significant improvement over assuming tactic latency and tactic cost to be static values.}

\section{Related Works} 
\label{sec: relatedworks}

Although our work is, to the best of our knowledge, the first known to make tactic volatility a first-class concern in the self-adaptive decision-making process, previous works have examined the impact of tactic latency in self-adaptive systems. C{\'a}mara \etal\cite{camara2014stochastic} was likely the first to consider tactic latency and examined how considering latency could be used to assist the proactive adaptation process. However, this work differs from TVA in that it does not consider any forms of tactic volatility that are likely to be encountered.

In SB-PLA, the latency of an adaptation is considered in the adaptation decision-making process~\cite{kwiatkowska2004probabilistic}. A primary benefit of SB-PLA is that systems that cannot use tactic-based adaptations can still include latency awareness in their decision-making process~\cite{moreno2017adaptation}. In addition to supporting pro-activeness and concurrent tactic execution, PLA techniques also account for latency. PLA considers the amount of time necessary for a tactic to execute, in order to avoid situations that are not achievable when time dimensions are recognized. However, like with many latency aware approaches, latency is still considered to be a static attribute. In our work, we do not consider latency to be static and provide the system with the ability to predict tactic latency at run time.

Jamshidi \etal\cite{Jamshidi:2015:SCC:2861850.2861987} presented \texttt{FQL4KE}, a self-learning fuzzy cloud controller. This enables systems to not rely upon design-time knowledge, but allows users to simply adjust weights that represent system priorities. This work found that their proposed process outperformed a previously devised technique that did not have a learning mechanism. Our work is similar in that we both utilize learning to enable the system to make better decisions. However, our work differs in that Jamshidi \etal focused on improving resource planning, and not in addressing tactic volatility as is accomplished by TVA.

While our work is the first to account for \emph{cost volatility}, existing research has considered cost in the self-adaptive decision-making process. Several works have included cost in their utility equations, however they consider it to be a static value and do not account for real-world volatility~\cite{Moreno:2018:FED:3208359.3149180, 7573126, moreno2017adaptation, moreno2015proactive}. Jung~\etal\cite{Jung:2009:CAE:1813355.1813367} demonstrated that ignoring cost can have a significant impact on the ability to satisfy response-time-based SLAs. This work also proposed a cost-sensitive self-adaption engine using middleware to create adaptation decisions. This work differs from ours in that it only considers cost in cloud-based controllers, while we focus on cost during the entire decision-making process.

Esfahani \etal\cite{esfahani2013learning} utilized learning to improve the self-adaptive process. A primary contribution of this work is a new process of reasoning and assessing adaptation decisions using online learning. A preliminary work by Elkhodary~\cite{7503491} proposed combining feature-orientation, learning and dynamic optimization techniques to create a new class of self-adaptive systems that would be able to modify their adaptation logic at run time. TVA differs from this work in that learning is utilized to predict tactic latency and cost at run time, while also providing away to estimate future values for requirements. 

Kinneer \etal\cite{kinneer2019information} developed a process for reusing prior planning knowledge to help the system to adapt to unexpected situations. This process considered that tactics may fail, and supports reasoning about tactic latency. While this work is helpful for assisting the overall planning process, it does not enable the system to actively learn and predict future values for tactic latency and cost like our TVA approach does.



Machine learning has been utilized to help determine the most efficient configurations for self-adaptive systems, while also performing adaptation planning. Quin \etal\cite{Quin:2019:EAL:3341527.3341529} enhanced the traditional MAPE-K feedback loop through the use of a learning model that selects subsets of adaptation options from a larger set of adaptation possibilities. This process enables the system to make more efficient analysis decisions. Jamshidi \etal\cite{Jamshidi:2019:MLM:3341527.3341534} used machine learning to discover Pareto-optimal configurations to eliminate the need to explore every configuration. This work also restricted the search space necessary to make planning tractable. These works differ from ours in that while they use machine learning to create a more efficient self-adaptive process, they do not use machine learning to address the issue of tactic volatility.



The popular FIFA 98 dataset~\cite{arlitt2000workload, ita.ee.lbl.gov_Fifa_URL} is a collection of requests made to the 1998 World Cup website over an approximately four month span. This dataset is widely used in self-adaptive research and can be used to simulate when new servers would have to be activated up to handle additional web traffic or when to disable resource heavy features on a website because the traffic load is impeding the system. However, this dataset does not contain latency. Each request is a log entry that consists of information such as timestamp, size, status, and URL hit. Because this dataset does not contain any latency information, such as a request received and request fulfillment time, it cannot be used in collaboration with researching and evaluating latency-aware strategies such as our proposed TVA process.



\section{Threats and Future Work}
\label{sec: threats}

Our evaluations have demonstrated TVA's ability to enable systems to better account for tactic volatility. However, there are limitations to this work. In many systems, tactic cost may be an ambiguous and tough-to-define measurement. This inability to accurately measure cost could inhibit the adoption of our process by limiting the quality and quantity of observed input values into our prediction process. Furthermore, cost can also be a relative term, and in our TVA approach we consider it to be a quantifiable value. For example, one could argue that the `cost' for performing an action could be the wear and tear on a physical component in the device. Such cost is difficult to quantify in most situations. Therefore, when using TVA the notion of cost must be restricted to a value that is reasonably easy to quantify.

Although TVA provides a method of monitoring specifications defined in the SLA, not all specifications are necessarily measurable with time-series analysis. For example, a system may have the specification that it must be available 99.99\% of the time. This is not something that could be measured or predicted using time-series analysis, rather it would need to be accounted for using fault detection techniques in the system's architecture. In systems that do not directly utilize a SLA, the improved tactic estimations can still be used to benefit the decision-making process by providing more informed and accurate tactic attribute values. Future work should be conducted to examine precisely how our TVA process can be incorporated into these systems and determine the benefits that they will have.



To be completely proactive, future work must be done to update our ARIMA method to consider tactic latency. Currently, this process can alert the system of a specification that is about to be broken, however the time between monitoring intervals may not leave enough time to fully execute a tactic. Therefore, there may be occurrences where specifications are slightly broken for a period of time. This future work will likely examine other time series models as well that can be flexible to different systems and datasets. This in turn would also help us to start addressing the problem of not having enough data to build other kinds of models and would allow this work to develop into an entire decision-making framework.

Although we have demonstrated the benefits of our TVA approach with real-world experimental data, we have yet to implement TVA on any physical devices. Future work will consist of including our adaptation process into physical equipment such as IoT devices, small unmanned aerial vehicles (UAV), and self-adaptive web systems. 

ARIMA demonstrated its ability to perform time series forecasting, however no mechanisms are in place to quantify uncertainty within these predictions. Future work could be done to include confidence intervals around predictions made by ARIMA. For example, instead of only predicting a single point value for a specification, we could predict a range that states something such as the following, ``\textit{we are 95\% certain response time will be between 3.1 and 3.7 seconds}''. This would allow the system to have a ``buffer'' zone around its predictions, therefore providing the decision-making process with more information.

Additionally, the ARIMA models utilized in this work were pre-trained on gathered historical data. It is also possible to have the ARMIA models be updated online as new data is gathered by a system to adapt themselves as more information becomes available. Other methods for time series prediction such as recurrent neural networks (RNNs), which have been successfully used for a variety of time series forecasting problems~\cite{di2016artificial,maknickiene2012application,gers-lstm-forget-2000,felder2010wind,choi2015doctor,elsaid2018optimizing,elsaid2019evolving}, can be examined as well. RNNs may also potentially be able to incorporate non-time series data into predictions.


Our evaluation data was created using two Raspberry Pi's, and does not simulate a complicated system such as a self-driving car or a UAV. However, this generated data was intended to help demonstrate the capabilities of TVA and the benefits of accounting for tactic volatility. In reality, the data generated by these tools could represent virtually any form of tactic volatility (\eg the time required for a UAV to communicate with a base station or the energy cost of a self-driving car performing a tactic).

There are a few potential limitations to the use of our VALET tool and dataset in the evaluation of our proposed TVA process. First, VALET generates its data by performing a small number of tasks. A real-world self-adaptive system would likely perform a large number of tasks in any given adaptation, which depending on the system, could impact one another. VALET is also limited in the forms of variability that it may encounter as opposed to a real-world system. For example, VALET is significantly less likely to be actively targeted by human hackers than many real-world self-adaptive systems, which could limit the encountered variability. However, it is somewhat unreasonable to expect that any created evaluation system would have the capabilities to address a majority of real-world events and possibilities. Despite these possible limitations, we are confident in the ability of VALET in creating satisfactory evaluation data for not only our study, but for future research conducted by others.

\section{Conclusion}
\label{sec: conclusion}
In this work, we propose a Tactic Volatility Aware (TVA) process that is able to account for tactic volatility in multiple ways. TVA first uses time-series forecasting with an autoregressive integrated moving average (ARIMA) model to monitor system specifications defined in the SLA, supporting the system in acting more proactively while maintaining them. Next, TVA uses multiple regression analysis (MRA) to predict tactic latency and cost helping to improve the decision-making process. Our contribution also includes a tool that utilizes physical devices to create real-world tactic volatility data.



\section*{Acknowledgements}
Elements of this work are sponsored by the United States Air Force Research Laboratory (AFRL).

\balance
\clearpage 
\bibliographystyle{abbrv}
\bibliography{tactics}

\end{document}